\documentclass{article}
\usepackage{ijcai15}
\usepackage{multirow}
\usepackage{times}
\usepackage{url}
\usepackage{graphicx,epstopdf}
\usepackage{latexsym}
\usepackage[tight]{subfigure}
\usepackage{algorithm}
\usepackage{algorithmic}
\usepackage{amsmath} 
\usepackage{amssymb}  
\usepackage{cases}
\usepackage{float}
\usepackage{array}
\newcolumntype{C}[1]{>{\centering}p{#1}}

\DeclareMathOperator*{\argmin}{arg~min}

\title{Learning Embedding Representations for Knowledge Inference on Imperfect and Incomplete Repositories}
\author{Miao Fan$^\dagger$, Qiang Zhou and Thomas Fang Zheng$^\ddagger$ \\
CSLT, Division of Technical Innovation and Development\\
Tsinghua National Laboratory for Information Science and Technology\\ Department of Computer Science and Technology\\
 Tsinghua University, Beijing, 100084, China\\
$\dagger$fanmiao.cslt.thu@gmail.com,$\ddagger$fzheng@tsinghua.edu.cn}

\begin{document}

\maketitle

\begin{abstract}
This paper considers the problem of knowledge inference on large-scale {\it imperfect} repositories with {\it incomplete} coverage by means of embedding entities and relations at the first attempt. We propose {\it IIKE} (Imperfect and Incomplete Knowledge Embedding), a probabilistic model which measures the probability of each belief, i.e. $\langle h,r,t\rangle$, in large-scale knowledge bases such as NELL and Freebase, and our objective is to learn a better low-dimensional vector representation for each entity ($h$ and $t$) and relation ($r$) in the process of minimizing the loss of fitting the corresponding confidence given by machine learning (NELL) or crowdsouring (Freebase), so that we can use $||{\bf h} + {\bf r} - {\bf t}||$ to assess the plausibility of a belief when conducting inference. We use subsets of those inexact knowledge bases to train our model and test the performances of {\it link prediction} and {\it triplet classification} on ground truth beliefs, respectively. The results of extensive experiments show that {\it IIKE} achieves significant improvement compared with the baseline and state-of-the-art approaches.
\end{abstract}

\section{Introduction}
The explosive growth in the number of web pages has drawn much attention to the study of information extraction \cite{sarawagi2008information} in recent decades. The aim of this is to distill unstructured online texts, so that we can store and exploit the distilled information as structured knowledge.
Thanks to the long-term efforts made by experts, crowdsouring and even machine learning techniques, several web-scale knowledge repositories have been built, such as Wordnet\footnote{\url{http://wordnet.princeton.edu/}}, Freebase\footnote{\url{https://www.freebase.com/}} and NELL\footnote{\url{http://rtw.ml.cmu.edu/rtw/}}, and most of them contain tens of millions of extracted beliefs which are commonly represented by triplets, i.e. $\langle head\_entity, relation, tail\_entity\rangle$.

Although we have gathered colossal quantities of beliefs, state of the art work in the literature \cite{42024} reported that in this field, our knowledge bases are far from complete. For instance, nearly 97\% persons in Freebase have unknown parents.
To populate incomplete knowledge repositories, a large proportion of researchers follow the classical approach by extracting knowledge from texts \cite{zhou-EtAl:2005:ACL,bach2007review,mintz2009distant}. For example, they explore ideal approaches that can automatically generate a precise belief like $\langle Madrid, capital\_city\_of, Spain\rangle$ from the sentence ``{\it Madrid is the capital and largest city of Spain.}''\footnote{\url{http://en.wikipedia.org/wiki/Madrid}} on the web. However, even cutting-edge research \cite{fan-EtAl:2014:P14-1} could not satisfy the demand of web-scale deployment, due to the diversification of natural language expression. Moreover, many implicit relations between two entities which are not recorded by web texts still need to be mined.

Therefore, some recent studies focus on inferring undiscovered beliefs based on the knowledge base itself without using extra web texts. One representative idea is to consider the whole repository as a graph where entities are nodes and relations are edges. The canonical approaches \cite{Quinlan:1993:FMR:645323.649599,lao2010relational,lao-mitchell-cohen:2011:EMNLP,conf/emnlp/GardnerTKM13} generally conduct {\it relation-specific} random walk inference based on the {\it local connectivity patterns} learnt from the imperfect knowledge graph.
An alternative paradigm aims to perform {\it open-relation} inference via embedding all the elements, including entities and relations, into low-dimensional vector spaces. The proposed methods \cite{Sutskever2009,Jenatton2012,Bordes2011,Bordes2013a,Socher2013,DBLP:conf/aaai/WangZFC14} show promising performance, however, by means of learning from ground-truth training knowledge.

This paper thus contributes a probabilistic knowledge embedding model called {\it IIKE}\footnote{short for Imperfect and Incomplete Knowledge Embedding.} to measure the probability of each triplet, i.e. $\langle h, r, t\rangle$, and our objective is to learn a better low-dimensional vector representation for each entity ($h$ and $t$) and relation ($r$) in the process of minimizing the loss of fitting the corresponding confidence given by machine learning (NELL) or crowdsouring (Freebase). To the best of our knowledge, {\it IIKE} is the first approach that attempts to learn {\it global connectivity patterns} for {\it open-relation} inference on imperfect and incomplete knowledge bases. In order to prove the effectiveness of the model, we conduct experiments on two tasks involved in knowledge inference, {\it link prediction} and {\it triplet classification}, using the two repositories mentioned above. Inexact beliefs are used to train our model, and we test the performance on ground truth beliefs. Results show that {\it IIKE} outperforms the other cutting-edge approaches on both different types of knowledge bases.

\section{Related Work}
We group recent research work related to self-inferring new beliefs based on knowledge repositories without extra texts into two categories, graph-based inference models \cite{Quinlan:1993:FMR:645323.649599,lao2010relational,lao-mitchell-cohen:2011:EMNLP,conf/emnlp/GardnerTKM13} and embedding-based inference models \cite{Sutskever2009,Jenatton2012,Bordes2011,Bordes2013a,Socher2013}, and describe the principal differences between them,
\begin{itemize}
  \item {\it Symbolic representation v.s. Distributed representation}: Graph-based models regard the entities and relations as atomic elements, and represent them in a symbolic framework. In contrast, embedding-based models explore distributed representations via learning a low-dimensional continuous vector representation for each entity and relation.

  \item {\it Relation-specific v.s. Open-relation}: Graph-based models aim to induce rules or paths for a specific relation first, and then infer corresponding new beliefs. On the other hand, embedding-based models encode all relations into the same embedding space and conduct inference without any restriction on some specific relation.
\end{itemize}

\subsection{Graph-based Inference}
Graph-based inference models generally learn the representation for specific relations from the knowledge graph.

{\it N-FOIL} \cite{Quinlan:1993:FMR:645323.649599} learns first order Horn clause rules to infer new beliefs from the known ones. So far, it has helped to learn approximately 600 such rules. However, its ability to perform inference over large-scale knowledge repositories is currently still very limited.

{\it PRA} \cite{lao2010relational,lao-mitchell-cohen:2011:EMNLP,conf/emnlp/GardnerTKM13} is a data-driven random walk model which follows the paths from the head entity to the tail entity on the local graph structure to generate non-linear feature combinations representing the labeled relation, and uses logistic regression to select the significant features which contribute to classifying other entity pairs belonging to the given relation.

\subsection{Embedding-based Inference}
Embedding-based inference models usually design various scoring functions $f_r(h, t)$ to measure the plausibility of a triplet $\langle h, r, t \rangle$. The lower the dissimilarity of the scoring function $f_r(h, t)$ is, the higher the compatibility of the triplet will be.

{\it Unstructured} \cite{Bordes2013a} is a naive model which exploits the occurrence information of the head and the tail entities without considering the relation between them. It defines a scoring function $||{\bf h}-{\bf t}||$, and this model obviously can not discriminate a pair of entities involving different relations. Therefore, {\it Unstructured} is commonly regarded as the baseline approach.

{\it Distance Model (SE)} \cite{Bordes2011} uses a pair of matrices $(W_{rh}, W_{rt})$, to characterize a relation $r$. The dissimilarity of a triplet is calculated by $||W_{rh}{\bf h} - W_{rt}{\bf t}||_1$. As pointed out by Socher et al. \shortcite{Socher2013}, the separating matrices $W_{rh}$ and $W_{rt}$ weaken the capability of capturing correlations between entities and corresponding relations, even though the model takes the relations into consideration.

{\it Single Layer Model}, proposed by Socher et al. \shortcite{Socher2013} thus aims to alleviate the shortcomings of the {\it Distance Model} by means of the nonlinearity of a single layer neural network $g(W_{rh}{\bf h} + W_{rt}{\bf t} + {\bf b}_r)$, in which $g = tanh$. The linear output layer then gives the scoring function: ${\bf u}^T_rg(W_{rh}{\bf h} + W_{rt}{\bf t} + {\bf b}_r)$.

{\it Bilinear Model} \cite{Sutskever2009,Jenatton2012} is another model that tries to fix the issue of weak interaction between the head and tail entities caused by {\it Distance Model} with a relation-specific bilinear form: $f_r(h, t) = {\bf h}^TW_r{\bf t}$.

{\it Neural Tensor Network (NTN)} \cite{Socher2013} designs a general scoring function:  $f_r(h, t) = {\bf u}^T_rg({\bf h}^TW_r{\bf t}+ W_{rh}{\bf h} + W_{rt}{\bf t} + {\bf b}_r)$, which combines the {\it Single Layer Model} and the {\it Bilinear Model}. This model is more expressive as the second-order correlations are also considered into the nonlinear transformation function, but the computational complexity is rather high.

{\it TransE} \cite{Bordes2013a} is a canonical model different from all the other prior arts, which embeds relations into the same vector space of entities by regarding the relation $r$ as a translation from $h$ to $t$, i.e. ${\bf h} + {\bf r} = {\bf t}$. It works well on the beliefs with ONE-TO-ONE mapping property but performs badly on multi-mapping beliefs. Given a series of facts associated with a ONE-TO-MANY relation $r$, e.g. ${(h, r, t_1), (h, r, t_2), ..., (h, r, t_m)}$, {\it TransE} tends to represent the embeddings of entities on MANY-side extremely the same with each other and hardly to be discriminated.

{\it TransH} \cite{DBLP:conf/aaai/WangZFC14} is the state of the art approach as far as we know. It improves {\it TransE} by modeling a relation as a hyperplane, which makes it more flexible with regard to modeling beliefs with multi-mapping properties.

Even though the prior arts of knowledge embedding are promising when conducting {\it open-relation} inference on large-scale bases, the stage they stand on is made of ground-truth beliefs. The model {\it IIKE} that we have proposed belongs to the embedding-based community, but firstly tackles the problem with knowledge inference based on imperfect and incomplete repositories. Nevertheless, we compare our approach with the methods mentioned above, and assess the performance with both the dataset and the metrics they have used as part of the extensive experiments.

\section{Model}
\begin{figure*}
\centering
\includegraphics[width=0.72\textwidth]{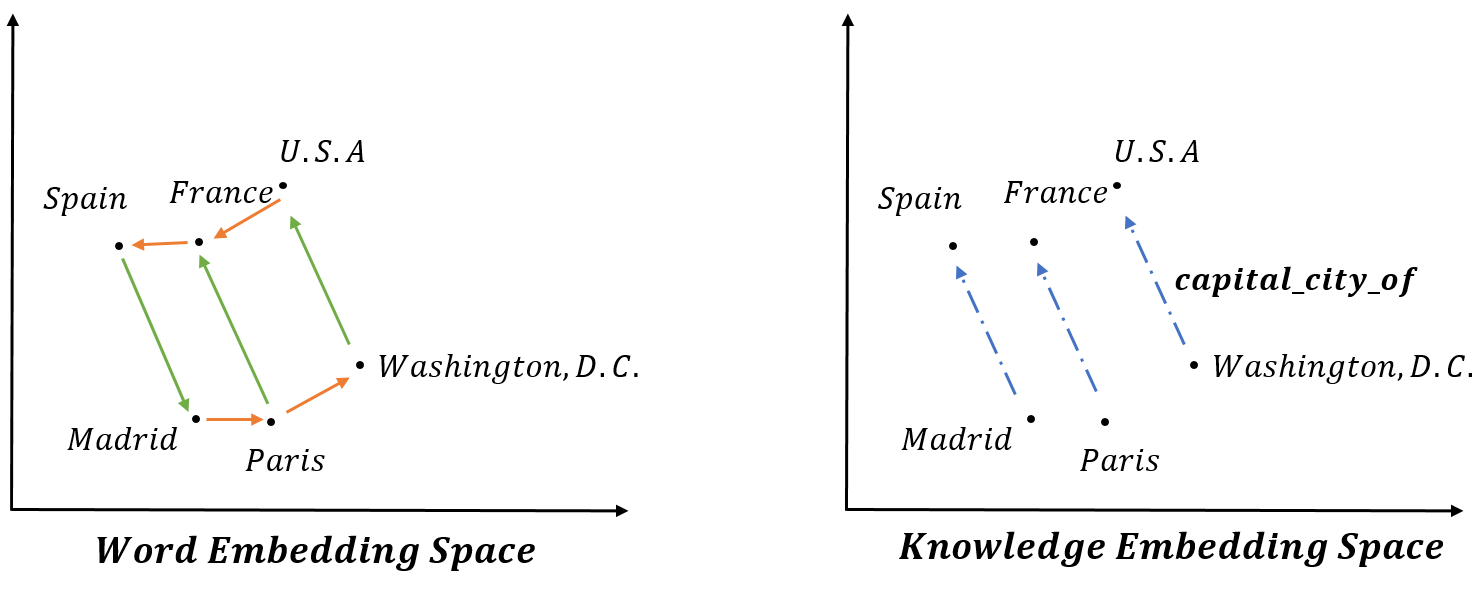}\\

\caption{The result of vector calculation in the word embedding space: ${\bf v}_{Madrid} - {\bf v}_{Spain} + {\bf v}_{France} \approx {\bf v}_{Paris}$ and ${\bf v}_{Madrid} - {\bf v}_{Spain} + {\bf v}_{U.S.A} \approx {\bf v}_{Washington, D.C.}$
The most possible reason of ${\bf v}_{Spain} - {\bf v}_{Madrid} \approx {\bf v}_{France} - {\bf v}_{Paris}$ and ${\bf v}_{Spain} - {\bf v}_{Madrid} \approx {\bf v}_{U.S.A} - {\bf v}_{Washington, D.C.}$, is that $capital\_city\_of$ is the shared relation. In other words, ${\bf h}_{Madrid} + {\bf r}_{capital\_city\_of} \approx {\bf t}_{Spain}$, if the belief $\langle Madrid, capital\_city\_of, Spain\rangle$ is plausible.}

\end{figure*}
The plausibility of a belief $\langle h, r, t \rangle$ can be regarded as the joint probability of the head entity $h$, the relation $r$ and the tail entity $t$, namely $Pr(h,r,t)$. Similarly, $Pr(h|r,t)$ stands for the conditional probability of predicting $h$ given $r$ and $t$. We assume that $Pr(h,r,t)$ is collaboratively influenced by $Pr(h|r,t)$, $Pr(r|h,t)$ and $Pr(t|h,r)$, and more specifically it equals to the geometric mean of $Pr(h|r,t)$$Pr(r|h,t)$$Pr(t|h,r)$, which is shown in the subsequent equation,
\begin{equation}
Pr(h,r,t) = \sqrt[3]{Pr(h|r,t)Pr(r|h,t)Pr(t|h,r)}.
\end{equation}

Given $r$ and $t$, there are multiple choices of $h'$ which may appear as the head entity. Therefore, if we use $E_h$ to denote the set of all the possible head entities given $r$ and $t$, $Pr(h|r,t)$ can be defined as
\begin{equation}
Pr(h|r,t) = \frac{\exp^{D(h,r,t)}}{\sum_{h' \in E_h}{\exp^{D(h',r,t)}}},
\end{equation}
The other factors, i.e. $Pr(r|h,t)$ and $Pr(r|h,t)$, are defined accordingly by slightly revising the normalization terms as shown in Equation (3) and (4), in which $R$ and $E_t$ represents the set of relations and tail entities, respectively.
\begin{equation}
Pr(r|h,t) = \frac{\exp^{D(h,r,t)}}{\sum_{r' \in R}{\exp^{D(h,r',t)}}}.
\end{equation}
\begin{equation}
Pr(t|h,r) = \frac{\exp^{D(h,r,t)}}{\sum_{t' \in E_t}{\exp^{D(h,r,t')}}}.
\end{equation}

The last function that we do not explain in Equation (2), (3) and (4) is $D(h,r,t)$. Inspired by somewhat surprising patterns learnt from word embeddings \cite{conf/naacl/MikolovYZ13} illustrated by Figure 1, the result of word vector calculation, for instance ${\bf v}_{Madrid} - {\bf v}_{Spain} + {\bf v}_{France}$, is closer to ${\bf v}_{Paris}$ than to any other words \cite{mikolov2013distributed}.
If we study the example mentioned above, the most possible reason ${\bf v}_{Spain} - {\bf v}_{Madrid} \approx {\bf v}_{France} - {\bf v}_{Paris}$, is that $capital\_city\_of$ is the relation between $Madrid$ and $Spain$ , and so is $Paris$ and $France$. In other words, ${\bf h}_{Madrid} + {\bf r}_{capital\_city\_of} \approx {\bf t}_{Spain}$, if the belief is plausible.
Therefore, we define $D(h,r,t)$ as follows to calculate the dissimilarity between ${\bf h} + {\bf r}$ and ${\bf t}$ using $L_1$ or $L_2$ norm, and set $b$ as the bias parameter.
\begin{equation}
D(h,r,t) = - ||{\bf h} + {\bf r} - {\bf t}|| + b.
\end{equation}

So far, we have already modeled the probability of a belief, i.e. $Pr(h,r,t)$. On the other hand, some imperfect repositories, such as NELL, which is automatically built by machine learning techniques \cite{carlson-aaai}, assign a confidence score ($[0.5-1.0]$) to evaluate the plausibility of the corresponding belief. Therefore, we define the cost function $\mathcal{L}$ shown in Equation (6), and our objective is to learn a better low-dimensional vector representation for each entity and relation while continuously minimizing the total loss of fitting each belief $\langle h,r,t,c \rangle$ in the training set $\Delta$ to the corresponding confidence $c$.
\begin{equation}
\begin{split}
& \argmin_{h,r,t}~~ \mathcal{L} = \sum_{\langle h,r,t,c \rangle \in \Delta}\frac{1}{2}(\log Pr(h,r,t) - \log c)^2\\
& ~~ = \sum_{\langle h,r,t,c\rangle \in \Delta} \frac{1}{2}\{\frac{1}{3}[ \log Pr(h|r,t) + \log Pr(r|h,t)\\
& ~~ + \log Pr(t|h,r)] - \log c)\}^2.
\end{split}
\end{equation}
\section{Algorithm}
To search for the optimal solution of Equation (6), we use {\it Stochastic Gradient Descent} (SGD) to update the embeddings of entities and relations in iterative fashion. However, it is cost intensive to directly compute the normalization terms in $Pr(h|r,t)$, $Pr(r|h,t)$ and $Pr(t|h,r)$. Enlightened by the work of Mikolov et al. \shortcite{mikolov2013distributed}, we have found an efficient approach that adopts negative sampling to transform the conditional probability functions, i.e. Equation (2), (3) and (4), to the binary classification problem, as shown in the subsequent equations,
\begin{equation}
\begin{split}
& \log Pr(h|r,t) \approx \log Pr(1|h,r,t) \\
& + \sum^k_{i = 1}{\mathbb{E}_{h'_i
\sim Pr(h' \in E_h)}\log Pr(0|h'_i, r, t)},
\end{split}
\end{equation}
\begin{equation}
\begin{split}
& \log Pr(r|h,t) \approx \log Pr(1|h,r,t) \\
& + \sum^k_{i = 1}{\mathbb{E}_{r'_i
\sim Pr(r' \in R)}\log Pr(0|h, r'_i, t)},
\end{split}
\end{equation}
\begin{equation}
\begin{split}
& \log Pr(t|h,r) \approx \log Pr(1|h,r,t) \\
& + \sum^k_{i = 1}{\mathbb{E}_{t'_i
\sim Pr(t' \in E_t)}\log Pr(0|h, r, t'_i)},
\end{split}
\end{equation}
in (7), (8), and (9), we sample $k$ negative beliefs and discriminate them from the positive case. For the simple binary classification problem mentioned above, we choose the logistic function with the offset $\epsilon$ shown in Equation (10) to estimate the probability that the given belief $\langle h, r, t\rangle$ is correct:
\begin{equation}
Pr(1|h,r,t) = \frac{1}{1 + \exp^{-D(h,r,t)}} + \epsilon.
\end{equation}

We also display the framework of the learning algorithm of {\it IIKE} in pseudocode as follows,

\begin{algorithm}
\caption{~~~~~The Learning Algorithm of {\bf IIKE}}
\begin{algorithmic}[1]
\REQUIRE ~~\\
Training set $\Delta = \{(h, r, t, c)\}$, entity set $E$, relation set $R$;
dimension of embeddings $d$, number of negative samples $k$, learning rate $\alpha$, convergence threshold $\eta$, maximum epoches $n$.\\
\STATE /*Initialization*/
\FOR{${\bf r} \in R$}
\STATE ${\bf r} := \text {Uniform} (\frac{-6}{\sqrt{d}}, \frac{6}{\sqrt{d}})$
\STATE ${\bf r} :=  \frac{{\bf r}}{|{\bf r}|} $\
\ENDFOR
\FOR{${\bf e} \in E$}
\STATE ${\bf e} := \text {Uniform}(\frac{-6}{\sqrt{d}}, \frac{6}{\sqrt{d}})$
\STATE ${\bf e} :=  \frac{{\bf e}}{|{\bf e}|} $\
\ENDFOR
\STATE /*Training*/
\STATE $i := 0$
\WHILE{$Rel. loss > \eta$ and $i < n$ }
\FOR{$\langle h, r, t \rangle \in \Delta$}
\FOR {$j \in$ range($k$)}
\STATE Negative sampling: $\langle h'_j, r, t \rangle \in \Delta'_h$
\STATE /*$\Delta'_h$ is the set of  $k$ negative beliefs replacing $h$*/
\STATE Negative sampling: $\langle h, r'_j, t \rangle \in \Delta'_r$
\STATE /*$\Delta'_r$ is the set of  $k$ negative beliefs replacing $r$*/
\STATE Negative sampling: $\langle h, r, t'_j \rangle \in \Delta'_t$
\STATE /*$\Delta'_t$ is the set of  $k$ negative beliefs replacing $t$*/
\ENDFOR
\STATE $\sum_{h, r, t, h', r', t'}\nabla \frac{1}{2}(\log Pr(h,r,t) - \log c)^2$
\STATE /*Updating embeddings of $\langle h, r, t \rangle \in \Delta, \langle h', r, t \rangle \in \Delta'_h, \langle h, r', t \rangle \in \Delta'_r, \langle h, r, t
'\rangle \in \Delta'_t$ with $\alpha$ and the batch gradients derived from Equation (7), (8), (9) and (10).*/
\ENDFOR
\STATE $i$++
\ENDWHILE
\\
\ENSURE ~~\\
All the embeddings of $h, t$ and $r$, where $h, t \in E$ and $ r \in R$.
\end{algorithmic}
\end{algorithm}

\section{Experiments}
Embedding the entities and relations into low-dimensional vector spaces facilitates several classical knowledge inference tasks, such as {\it link prediction} and {\it triplet classification}. More specifically, link prediction performs inference via predicting a ranked list of missing entities or relations given the other two elements of a triplet. For example, it can predict a series of $t$ given $h$ and $r$, or a bunch of $h$ given $r$ and $t$. And triplet classification is to discriminate whether a triplet $\langle h, r, t \rangle$ is correct or wrong.

Several recent research works \cite{Bordes2013a,Socher2013,DBLP:conf/aaai/WangZFC14} reported that they used subsets of Freebase ({\bf FB}) data to evaluate their models and showed the performance on the above two tasks, respectively. In order to conduct solid experiments, we compare our model ({\it IIKE}) with many related studies including the baseline and cutting-edge approaches mentioned in Section 2.2. Moreover, we use a larger imperfect and incomplete dataset ({\bf NELL}) to perform comparisons involving the same tasks to show the superior inference capability of {\it IIKE}, and have released this dataset for others to use.

We are also glad to share all the datasets, the source codes and the learnt embeddings for entities and relations, which can be freely downloaded from {\url{http://pan.baidu.com/s/1mgxGbg8}}.

\subsection{Link prediction}
One of the benefits of knowledge embedding is that we can apply simple vector calculations to many reasoning tasks, and link prediction is a valuable task that contributes to completing the knowledge graph.
With the help of knowledge embeddings, if we would like to tell whether the entity $h$ has the relation $r$ with the entity $t$, we just need to calculate the distance between ${\bf h + r}$ and ${\bf t}$. The closer they are, the more possibility the triplet $\langle h, r, t \rangle$ exists.
\subsubsection{Datasets}
\begin{table}[!htp]
\centering
\begin{tabular}{|c|c|c|c|}
  \hline
  {\bf DATASET} & {\bf FB15K} & {\bf NELL} \\
  \hline
  \hline
  \#(ENTITIES) & 14,951 & 74,037  \\
  \#(RELATIONS) &1,345 & 226    \\
  \#(TRAINING EX.) & 483,142 & 713,913\\
  \#(VALIDATING EX.) & 50,000 & 7,296 \\
  \#(TESTING EX.) & 59,071 & 7,296 \\
  \hline
\end{tabular}
\caption{Statistics of the datasets used for link prediction task.}
\end{table}
\begin{table*}[!htp]
  \centering
  \begin{tabular}{*{5}{|c|}}
    \hline
    {\bf DATASET} & \multicolumn{4}{|c|}{\bf FB15K}\\
    \hline
    \hline
    {\multirow{2}*{\bf METRIC}}
    &  \multicolumn{2}{|c|}{\em MEAN RANK} & \multicolumn{2}{|c|}{\em MEAN HIT@10}\\
   &  {\em Raw} & {\em Filter} & {\em Raw} & {\em Filter}\\
   \hline
   \hline
   {\bf Unstructured} \cite{Bordes2014} & 1,074 / 14,951 & 979 / 14,951& 4.5\% & 6.3\% \\
   {\bf RESCAL} \cite{Nickel2011} & 828 / 14,951 & 683 / 14,951& 28.4\% & 44.1\% \\
   {\bf SE} \cite{Bordes2011} & 273 / 14,951& 162 / 14,951& 28.8\% & 39.8\% \\
   {\bf SME (LINEAR)} \cite{Bordes2014} & 274 / 14,951& 154 / 14,951& 30.7\% & 40.8\% \\
   {\bf SME (BILINEAR)} \cite{Bordes2014} & 284 / 14,951& 158 / 14,951& 31.3\% & 41.3\% \\
   {\bf LFM} \cite{Jenatton2012}& 283 / 14,951& 164 / 14,951 & 26.0\% & 33.1\% \\
   {\bf TransE} \cite{Bordes2013a} & 243 / 14,951& 125 / 14,951& 34.9\% & 47.1\%  \\
   {\bf TransH} \cite{DBLP:conf/aaai/WangZFC14} & 211 / 14,951& 84 / 14,951& 42.5\% & 58.5\%  \\
   \hline
   {\bf IIKE}   & {\bf 183} / 14,951 & {\bf 70} / 14,951& {\bf 47.1\%} & {\bf 59.7\%} \\
   \hline
  \end{tabular}
  \caption{Link prediction results on the {\bf FB15K} dataset. We compared our proposed {\it IIKE} with the state-of-the-art method {\it TransH} and other prior arts mentioned in Section 2.2.}
\end{table*}

\begin{table*}[!htp]
  \centering
  \begin{tabular}{*{5}{|c|}}
    \hline
    {\bf DATASET} & \multicolumn{4}{|c|}{\bf NELL }\\
    \hline
    \hline
    {\multirow{2}*{\bf METRIC}}
    &  \multicolumn{2}{|c|}{\em MEAN RANK} & \multicolumn{2}{|c|}{\em MEAN HIT@10}\\
   &  {\em Raw} & {\em Filter} & {\em Raw} & {\em Filter}\\
   \hline
   \hline

   {\bf TransE} \cite{Bordes2013a} & 4,254 / 74,037 & 4,218 /  74,037 & 11.0\% & 12.3\% \\ 
   {\bf TransH} \cite{DBLP:conf/aaai/WangZFC14} & 3,469 /  74,037& {\bf 2,218} /  74,037 & 25.2\% & {\bf 41.6}\%  \\
   \hline

   {\bf IIKE}   & {\bf 2,464} / 74,037 & 2,428 / 74,037 & {\bf 37.3\%} &  38.2\% \\
   \hline
  \end{tabular}
  \caption{Link prediction results on the {\bf NELL} dataset. We compared our proposed {\it IIKE} with the cutting-edge methods {\it TransH} and {\it TransE}.}
\end{table*}

Bordes et al. \cite{Bordes2013a} released a large dataset ({\bf FB15K})\footnote{Related studies on this dataset can be looked up from the website \url{https://www.hds.utc.fr/everest/doku.php?id=en:transe}}, extracted from Freebase and constructed by crowdsourcing, in which each belief is a triplet without a confidence score. Therefore, we assign 1.0 to each training triplet by default. We have also identified a larger repository on the web named {\bf NELL}\footnote{The whole dataset of NELL can be downloaded from \url{http://rtw.ml.cmu.edu/rtw/resources}} which is automatically built by machine learning techniques, and each triplet is labeled with a probability estimated by synthetic algorithms \cite{carlson-aaai}. We reserve the beliefs with probability ranging (0.5 - 1.0], use the ground-truth (1.0) beliefs as the validating and testing examples, and train the models with the remains.

Table 1 shows the statistics of these two datasets. The scale of {\bf NELL} dataset is larger than {\bf FB15K} with many more entities but fewer relations, which may lead to the differences of tuning parameters\footnote{It turns out that embedding models prefer a larger dimension of vector representations for the dataset with more entities, and $ L_1$ {\it norm} for fewer relations.}.

\subsubsection{Evaluation Protocol}
For each testing triplet, all the other entities that appear in the training set take turns to replace the head entity. Then we get a bunch of candidate triplets associated with the testing triplet. The dissimilarity of each candidate triplet is firstly computed by various scoring functions, such as $||{\bf h + r - t}||$, and then sorted in ascending order. Finally, we locate the ground-truth triplet and record its rank. This whole procedure runs in the same way when replacing the tail entity, so that we can gain the mean results. We use two metrics, i.e. {\it Mean Rank} and {\it Mean Hit@10} (the proportion of ground truth triplets that rank in Top 10), to measure the performance. However, the results measured by those metrics are relatively inaccurate, as the procedure above tends to generate false negative triplets. In other words, some of the candidate triplets rank rather higher than the ground truth triplet just because they also appear in the training set. We thus filter out those triplets to report more reasonable results.
\subsubsection{Experimental Results}
We compared {\it IIKE} with the state-of-the-art {\it TransH}, {\it TransE} and other models mentioned in Section 2.2 evaluated on {\bf FB15K} and {\bf NELL }. We tuned the parameters of each previous model\footnote{All the codes for the related models can be downloaded from \url{https://github.com/glorotxa/SME}} based on the validation set, and select the combination of parameters which leads to the best performance. The results of prior arts on {\bf FB15K} are the same as those reported by Wang et al. \shortcite{DBLP:conf/aaai/WangZFC14}. For {\it IIKE}, we tried several combinations of parameters: $d = \{20, 50, 100\}$, $\alpha = \{0.1, 0.05, 0.01, 0.005,0.002\}$, $b = \{7.0, 10.0, 15.0\}$ and $norm = \{L_1, L_2\}$, and finally chose $d = 50$, $\alpha = 0.002$, $b = 7.0$, $norm = L_2$ for the {\bf FB15K} dataset, and $d = 100$, $\alpha = 0.001$, $b = 7.0$, $norm = L_1$ for the {\bf NELL} dataset. Moreover, to make responsible comparisons between {\it IIKE} and the state-of-the-art approaches, we requested the authors of {\it TransH}  to re-run their system on the {\bf NELL} dataset and reported the best results.  Table 2 demonstrates that {\it IIKE} outperforms all the prior arts, including the baseline model {\it Unstructured} \cite{Bordes2014}, {\it RESCAL} \cite{Nickel2011}, {\it SE} \cite{Bordes2011}, {\it SME (LINEAR)} \cite{Bordes2014}, {\it SME (BILINEAR)} \cite{Bordes2014}, {\it LFM} \cite{Jenatton2012} and {\it TransE} \cite{Bordes2013a}, and achieves significant improvements on the {\bf FB15K} dataset, compared with the state-of-the-art {\it TransH} \cite{DBLP:conf/aaai/WangZFC14}. For the {\bf NELL} dataset, {\it IIKE} performs stably on the evaluation metrics compared with {\it TransH} and {\it TransE}, as Table 3 shows that it improves by 28.9\% in terms of {\it Raw Mean Rank}, and achieves comparable performance of {\it Filter Mean Rank} compared with {\it TransH}.

\subsection{Triplet classification}
Triplet classification is another inference related task proposed by Socher et al. \shortcite{Socher2013} which focuses on searching a relation-specific threshold $\sigma_r$ to identify whether a triplet $\langle h, r, t \rangle$ is plausible.
\subsubsection{Datasets}
\begin{table}[!htp]
\centering
\begin{tabular}{|c|c|c|}
  \hline
  {\bf DATASET} & {\bf FB15K} & {\bf NELL}\\
  \hline
  \hline
  \#(ENTITIES) & 14,951 & 74,037\\
  \#(RELATIONS) & 1,345  & 226\\
  \#(TRAINING EX.)  & 483,142 & 713,913\\
  \#(VALIDATING EX.)  & 100,000 & 14,592\\
  \#(TESTING EX.) & 118,142 & 14,582 \\
  \hline
\end{tabular}
\caption{Statistics of the datasets used for triplet classification task.}
\end{table}
Wang et al. \shortcite{DBLP:conf/aaai/WangZFC14} constructed a standard dataset {\bf FB15K} sampled from Freebase. Moreover, we build another imperfect and incomplete dataset, i.e. {\bf NELL}, following the same principle that the head or the tail entity can be randomly replaced with another one to produce a negative triplet, but in order to build much tough validation and testing datasets, the principle emphasizes that the picked entity should once appear at the same position. For example, {\it (Pablo Picaso, nationality, American)} is a potential negative example rather than the obvious irrational {\it (Pablo Picaso, nationality, Van Gogh)}, given a positive triplet {\it (Pablo Picaso, nationality, Spanish)}, as {\it American} and {\it Spanish} are more common as the tails of {\it nationality}. And the beliefs in the training sets are the same as those used in triplet classification. Table 4 shows the statistics of the standard datasets that we used for evaluating models on the triplet classification task.

\subsubsection{Evaluation Protocol}
The decision strategy for binary classification is simple: if the dissimilarity of a testing triplet ($h, r, t$) computed by $f_r(h, t)$ is below the relation-specific threshold $\sigma_r$, it is predicted as positive, otherwise negative. The relation-specific threshold $\sigma_r$ can be searched via maximizing the classification accuracy on the validation triplets which belong to the relation $r$.
\subsubsection{Experimental Results}
\begin{table}
\centering
\begin{tabular}{|c|c|c|}
  \hline
  {\bf DATASET} & {\bf FB15K} & {\bf NELL}\\
  \hline
  \hline
  {\bf NTN} \cite{Socher2013}   & 66.7\%  & -\\
  {\bf TransE} \cite{Bordes2013a} & 79.7\%  & 82.4\%\\
  {\bf TransH} \cite{DBLP:conf/aaai/WangZFC14}  & 80.2\%  & 89.1\%\\
  \hline
  {\bf IIKE} & {\bf 91.1\%} & {\bf 91.4\%}\\
  \hline
\end{tabular}
  \caption{The accuracy of triplet classification compared with several latest approaches: {\it TransH}, {\it TransE} and {\it NTN}.}
\end{table}
We use the best combination of parameter settings in the link prediction task: $d = 50$, $\alpha = 0.002$, $b = 7.0$, $norm = L_2$ for the {\bf FB15K} dataset, and $d = 100$, $\alpha = 0.001$, $b = 7.0$, $norm = L_1$ for the {\bf NELL} dataset, to generate the entity and relation embeddings, and learn the best classification threshold $\sigma_r$ for each relation $r$. Compared with several of the latest approaches, i.e. {\it TransH} \cite{DBLP:conf/aaai/WangZFC14}, {\it TransE} \cite{Bordes2013a} and {\it Neural Tensor Network (NTN)}\footnote{Socher et al. reported higher classification accuracy in \cite{Socher2013} with word embeddings. In order to conduct a fair comparison, the accuracy of {\it NTN} reported in Table 5 is same with the EV (entity vectors) results in Figure 4 of \cite{Socher2013}.} \cite{Socher2013}, the proposed {\it IIKE} approach still outperforms them, as shown in Table 5. We also drew the precision-recall curves which indicate the capability of global discrimination by ranking the distance of all the testing triplets, and Figure 2 shows that the AUC (Areas Under the Curve) of {\it IIKE}  is much bigger than the other approaches.
\begin{figure}

\includegraphics[width=0.5\textwidth]{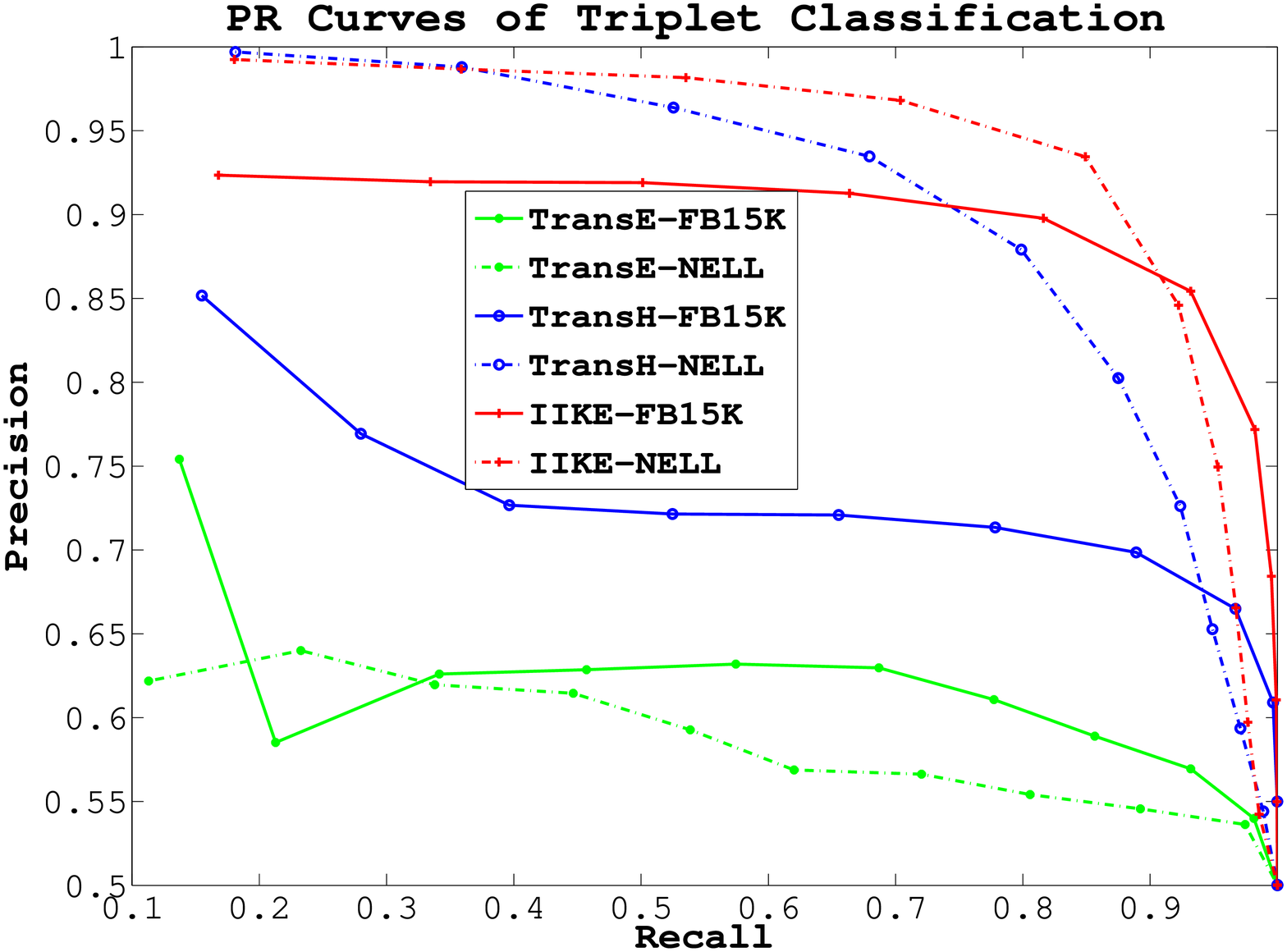}
\vspace{-10mm}
  \caption{The comparison of precison-recall curves for triplet classification among the proposed {\it IIKE} (red lines), the state-of-the-art approaches {\it TransH} (blue lines) and {\it TransE} (green lines).}
\end{figure}
\section{Conclusion}
We challenge the problem of knowledge inference on imperfect and incomplete repositories in this paper, and have produced an elegant probabilistic embedding model to tackle this issue at the first attempt by measuring the probability of a given belief $\langle h, r, t \rangle$. To efficiently learn the embeddings for each entity and relation, we also adopt the negative sampling technique to transform the original model and display the algorithm based on SGD to search the optimal solution. Extensive experiments on knowledge inference including {\it link prediction} and {\it triplet classification} show that our approach achieves significant improvement on two large-scale knowledge bases, compared with state-of-the-art and baseline methods.

We are pleased to see further improvements of the proposed model, which leaves open promising directions for the future work, such as taking advantage of the knowledge embeddings to enhance the studies of text summarization and open-domain question answering.

\section{Acknowledgments}
This work is supported by National Program on Key Basic Research Project (973 Program) under Grant 2013CB329304, National Science Foundation of China (NSFC) under Grant No.61373075.
\bibliographystyle{named}

\begin{thebibliography}{}

\bibitem[\protect\citeauthoryear{Bach and Badaskar}{2007}]{bach2007review}
Nguyen Bach and Sameer Badaskar.
\newblock A review of relation extraction.
\newblock {\em Literature review for Language and Statistics II}, 2007.

\bibitem[\protect\citeauthoryear{Bordes \bgroup \em et al.\egroup
  }{2011}]{Bordes2011}
Antoine Bordes, Jason Weston, Ronan Collobert, Yoshua Bengio, et~al.
\newblock Learning structured embeddings of knowledge bases.
\newblock In {\em AAAI}, 2011.

\bibitem[\protect\citeauthoryear{Bordes \bgroup \em et al.\egroup
  }{2013}]{Bordes2013a}
Antoine Bordes, Nicolas Usunier, Alberto Garcia-Duran, Jason Weston, and Oksana
  Yakhnenko.
\newblock Translating embeddings for modeling multi-relational data.
\newblock In {\em Advances in Neural Information Processing Systems}, pages
  2787--2795, 2013.

\bibitem[\protect\citeauthoryear{Bordes \bgroup \em et al.\egroup
  }{2014}]{Bordes2014}
Antoine Bordes, Xavier Glorot, Jason Weston, and Yoshua Bengio.
\newblock A semantic matching energy function for learning with
  multi-relational data.
\newblock {\em Machine Learning}, 94(2):233--259, 2014.

\bibitem[\protect\citeauthoryear{Carlson \bgroup \em et al.\egroup
  }{2010}]{carlson-aaai}
Andrew Carlson, Justin Betteridge, Bryan Kisiel, Burr Settles, Estevam
  R.~Hruschka Jr., and Tom~M. Mitchell.
\newblock Toward an architecture for never-ending language learning.
\newblock In {\em Proceedings of the Twenty-Fourth Conference on Artificial
  Intelligence (AAAI 2010)}, 2010.

\bibitem[\protect\citeauthoryear{Fan \bgroup \em et al.\egroup
  }{2014}]{fan-EtAl:2014:P14-1}
Miao Fan, Deli Zhao, Qiang Zhou, Zhiyuan Liu, Thomas~Fang Zheng, and Edward~Y.
  Chang.
\newblock Distant supervision for relation extraction with matrix completion.
\newblock In {\em Proceedings of the 52nd Annual Meeting of the Association for
  Computational Linguistics (Volume 1: Long Papers)}, pages 839--849,
  Baltimore, Maryland, June 2014. Association for Computational Linguistics.

\bibitem[\protect\citeauthoryear{Gardner \bgroup \em et al.\egroup
  }{2013}]{conf/emnlp/GardnerTKM13}
Matt Gardner, Partha~Pratim Talukdar, Bryan Kisiel, and Tom~M. Mitchell.
\newblock Improving learning and inference in a large knowledge-base using
  latent syntactic cues.
\newblock In {\em EMNLP}, pages 833--838. ACL, 2013.

\bibitem[\protect\citeauthoryear{Jenatton \bgroup \em et al.\egroup
  }{2012}]{Jenatton2012}
Rodolphe Jenatton, Nicolas Le~Roux, Antoine Bordes, Guillaume Obozinski, et~al.
\newblock A latent factor model for highly multi-relational data.
\newblock In {\em NIPS}, pages 3176--3184, 2012.

\bibitem[\protect\citeauthoryear{Lao and Cohen}{2010}]{lao2010relational}
Ni~Lao and William~W Cohen.
\newblock Relational retrieval using a combination of path-constrained random
  walks.
\newblock {\em Machine learning}, 81(1):53--67, 2010.

\bibitem[\protect\citeauthoryear{Lao \bgroup \em et al.\egroup
  }{2011}]{lao-mitchell-cohen:2011:EMNLP}
Ni~Lao, Tom Mitchell, and William~W. Cohen.
\newblock Random walk inference and learning in a large scale knowledge base.
\newblock In {\em Proceedings of the 2011 Conference on Empirical Methods in
  Natural Language Processing}, pages 529--539, Edinburgh, Scotland, UK., July
  2011. Association for Computational Linguistics.

\bibitem[\protect\citeauthoryear{Mikolov \bgroup \em et al.\egroup
  }{2013a}]{mikolov2013distributed}
Tomas Mikolov, Ilya Sutskever, Kai Chen, Greg~S Corrado, and Jeff Dean.
\newblock Distributed representations of words and phrases and their
  compositionality.
\newblock In C.J.C. Burges, L.~Bottou, M.~Welling, Z.~Ghahramani, and K.Q.
  Weinberger, editors, {\em Advances in Neural Information Processing Systems
  26}, pages 3111--3119. 2013.

\bibitem[\protect\citeauthoryear{Mikolov \bgroup \em et al.\egroup
  }{2013b}]{conf/naacl/MikolovYZ13}
Tomas Mikolov, Wen tau Yih, and Geoffrey Zweig.
\newblock Linguistic regularities in continuous space word representations.
\newblock In {\em HLT-NAACL}, pages 746--751. The Association for Computational
  Linguistics, 2013.

\bibitem[\protect\citeauthoryear{Mintz \bgroup \em et al.\egroup
  }{2009}]{mintz2009distant}
Mike Mintz, Steven Bills, Rion Snow, and Dan Jurafsky.
\newblock Distant supervision for relation extraction without labeled data.
\newblock In {\em Proceedings of the Joint Conference of the 47th Annual
  Meeting of the ACL and the 4th International Joint Conference on Natural
  Language Processing of the AFNLP: Volume 2-Volume 2}, pages 1003--1011.
  Association for Computational Linguistics, 2009.

\bibitem[\protect\citeauthoryear{Nickel \bgroup \em et al.\egroup
  }{2011}]{Nickel2011}
Maximilian Nickel, Volker Tresp, and Hans-Peter Kriegel.
\newblock A three-way model for collective learning on multi-relational data.
\newblock In {\em Proceedings of the 28th international conference on machine
  learning (ICML-11)}, pages 809--816, 2011.

\bibitem[\protect\citeauthoryear{Quinlan and
  Cameron-Jones}{1993}]{Quinlan:1993:FMR:645323.649599}
J.~Ross Quinlan and R.~Mike Cameron-Jones.
\newblock Foil: A midterm report.
\newblock In {\em Proceedings of the European Conference on Machine Learning},
  ECML '93, pages 3--20, London, UK, UK, 1993. Springer-Verlag.

\bibitem[\protect\citeauthoryear{Sarawagi}{2008}]{sarawagi2008information}
Sunita Sarawagi.
\newblock Information extraction.
\newblock {\em Foundations and trends in databases}, 1(3):261--377, 2008.

\bibitem[\protect\citeauthoryear{Socher \bgroup \em et al.\egroup
  }{2013}]{Socher2013}
Richard Socher, Danqi Chen, Christopher~D Manning, and Andrew Ng.
\newblock Reasoning with neural tensor networks for knowledge base completion.
\newblock In {\em Advances in Neural Information Processing Systems}, pages
  926--934, 2013.

\bibitem[\protect\citeauthoryear{Sutskever \bgroup \em et al.\egroup
  }{2009}]{Sutskever2009}
Ilya Sutskever, Ruslan Salakhutdinov, and Joshua~B Tenenbaum.
\newblock Modelling relational data using bayesian clustered tensor
  factorization.
\newblock In {\em NIPS}, pages 1821--1828, 2009.

\bibitem[\protect\citeauthoryear{Wang \bgroup \em et al.\egroup
  }{2014}]{DBLP:conf/aaai/WangZFC14}
Zhen Wang, Jianwen Zhang, Jianlin Feng, and Zheng Chen.
\newblock Knowledge graph embedding by translating on hyperplanes.
\newblock In {\em Proceedings of the Twenty-Eighth {AAAI} Conference on
  Artificial Intelligence, July 27 -31, 2014, Qu{\'{e}}bec City, Qu{\'{e}}bec,
  Canada.}, pages 1112--1119, 2014.

\bibitem[\protect\citeauthoryear{West \bgroup \em et al.\egroup }{2014}]{42024}
Robert West, Evgeniy Gabrilovich, Kevin Murphy, Shaohua Sun, Rahul Gupta, and
  Dekang Lin.
\newblock Knowledge base completion via search-based question answering.
\newblock In {\em WWW}, 2014.

\bibitem[\protect\citeauthoryear{Zhou \bgroup \em et al.\egroup
  }{2005}]{zhou-EtAl:2005:ACL}
GuoDong Zhou, Jian Su, Jie Zhang, and Min Zhang.
\newblock Exploring various knowledge in relation extraction.
\newblock In {\em Proceedings of the 43rd Annual Meeting of the Association for
  Computational Linguistics (ACL'05)}, pages 427--434, Ann Arbor, Michigan,
  June 2005. Association for Computational Linguistics.

\end{thebibliography}

\end{document}